\documentclass{article}
\usepackage{spconf,amsmath,graphicx}
\usepackage{graphics}
\usepackage[named]{algo}
\usepackage{stfloats}
\usepackage{amssymb}
\usepackage{amsmath}
\usepackage{amsfonts}
\usepackage{subfigure}
\usepackage{float}
\usepackage{pifont}
\usepackage{setspace}
\usepackage{url}
\usepackage{paralist}

\newcommand{\Section}{\section}
\newcommand{\SubSection}{\subsection}

\newtheorem{lemma}{Lemma}

\title{Robust Binary Fused Compressive Sensing using Adaptive Outlier Pursuit}
\name{Xiangrong Zeng\ \  and \  M\'{a}rio A. T. Figueiredo
\thanks{Work partially supported by Funda\c{c}\~{a}o para a Ci\^{e}ncia  e Tecnologia, grants PEst-OE/EEI/LA0008/2013 and PTDC/EEI-PRO/1470/2012. Xiangrong Zeng is partially supported by grant SFRH/BD/75991/2011.}}
\address{Instituto de Telecomunica\c{c}\~oes, Instituto Superior T\'ecnico, Universidade de Lisboa, Portugal}

\begin{document}
%
\maketitle
\begin{abstract}
We propose a new method, {\it robust binary fused compressive sensing} (RoBFCS), to recover sparse
piece-wise smooth signals from 1-bit compressive measurements.
The proposed method is a modification of our previous {\it binary fused compressive sensing} (BFCS)
algorithm, which is based on the {\it binary iterative hard thresholding} (BIHT) algorithm.
As in BIHT, the data term of the objective function is a one-sided $\ell_1$ (or $\ell_2$) norm.
Experiments show that the proposed algorithm is able to take advantage of the piece-wise
smoothness of the original signal and detect sign flips and correct them,
achieving more accurate recovery than BFCS and BIHT.
\end{abstract}
\begin{keywords}
1-bit compressive sensing, iterative hard thresholding, group sparsity, signal recovery.
\end{keywords}
\Section{Introduction}\label{sec:intro}

In {\it compressive sensing} (CS) \cite{candes2006stable}, \cite{donoho2006},
a sparse signal ${\bf x} \in {\mathbb R}^n$ is shown to be recoverable from a few  linear measurements
\begin{equation}\label{linearmodel}
 {\bf b}={\bf A}{\bf x},
\end{equation}
where  ${\bf b}\in\mathbb{R}^{m}$ (with $m<n$), ${\bf A}\in\mathbb{R}^{m\times n}$
is the sensing matrix (which satisfies some conditions), and the fact that $ m < n$
makes (\ref{linearmodel}) an ill-posed problem. This classical formulation assumes
real-valued measurements, thus ignoring that, in reality, any acquisition involves
quantization. In {\it quantized CS} (QCS) \cite{sun2009quantization},
 \cite{zymnis2010compressed}, \cite{laska2011democracy}, \cite{jacques2011dequantizing},
\cite{laska2012regime}, \cite{liu2012robust}, this fact is taken into account.
 An interesting extreme case of QCS is
1-bit CS \cite{boufounos20081},
\begin{equation}\label{1bitcs}
 {\bf y}=\mbox{sign}\left({\bf A}{\bf x}\right),
\end{equation}
where $\mbox{sign}(\cdot)$ is the element-wise sign function.
Such 1-bit measurements can be acquired
by a comparator with zero, which is very inexpensive and fast, as well as
robust to amplification distortions. In contrast with the measurement model of
standard CS, 1-bit measurements are blind to the magnitude of the original signal
${\bf x}$; the goal may then only be to recover ${\bf x}$, up to an intrinsically
unrecoverable magnitude.

The first algorithm for signal recovery from 1-bit measurements,
named {\it renormalized fixed point iteration} (RFPI) was proposed in \cite{boufounos20081}.
Soon after, \cite{boufounos2010reconstruction} showed that recovery from nonlinearly
distorted measurements is also possible, and \cite{boufounos2010reconstruction} proposed a
greedy algorithm ({\it matching sign pursuit}) \cite{boufounos2009greedy}. After
that seminal work, several algorithms for 1-bit CS have appeared;
a partial list includes linear programming  \cite{plan2011one}, \cite{plan2012robust},
{\it restricted-step shrinkage} \cite{laska2011trust}, and {\it binary iterative hard thresholding}
(BIHT), which performs better than the previous algorithms.
Algorithms for 1-bit CS, based on {\it generalized approximate message passing}
\cite{rangan2011generalized} and {\it majorization-minimization} \cite{hunter2004tutorial}, were proposed in \cite{fang2012fast} and \cite{kamilov2012one}, respectively. In \cite{shen2013blind}, the
$\ell_1$ norm in the data-term of \cite{plan2011one} was replaced by an $\ell_0$
norm; the resulting problem is solved by successive approximations,
yielding  a sequence of simpler problems, not requiring prior knowledge about the
sparsity of the original signal. Considering the possibility of sign flips due to by noise,
\cite{yan2012robust} introduced the {\it adaptive outlier pursuit} (AOP) algorithm,
and \cite{movahed2012robust} extended it into an algorithm termed {\it noise-adaptive RFPI},
which doesn't need prior information on the signal sparsity and number of sign flips.

More recently, \cite{bourquard2012binary} and \cite{yang2012bits} applied 1-bit CS in
image acquisition, \cite{davenport20121} studied matrix completion from noisy 1-bit
observations, \cite{xu2013statistical} used methods of statistical mechanics to
examine typical properties of 1-bit CS. The authors of \cite{bahmani2013robust} addressed
1-bit CS  using their recent {\it gradient support pursuit} (GraSP)
\cite{bahmani2011greedy} algorithm; finally, a {\it quantized iterative hard thresholding} method
proposed in \cite{jacques2013quantized} provides a bridge between 1-bit and high-resolution QCS.

Recently, we proposed {\it binary fused compressive sensing} (BFCS) \cite{zeng2012binary},
\cite{zeng2013D} to recover group-sparse signals from 1-bit CS measurements.
The rationale is that group-sparsity may express more structured knowledge about the
unknown signal than simple sparsity, thus potentially allowing for more robust recovery
from fewer measurements. In this paper, we further consider the possibility of sign
flips, and propose {\it robust BFCS} (RoBFCS) based on the AOP method
\cite{yan2012robust}.

 The rest of the paper is organized as follows: Section II reviews
the BIHT and BFCS algorithms, and introduces the proposed RoBFCS method;
Section III reports experimental results and Section IV concludes the paper.

\Section{Robust BFCS}
\SubSection{The Observation Model}
In this paper, we consider the noisy 1-bit measurement model,
\begin{equation}\label{1bitcsnoisy}
 {\bf y}=\mbox{sign}\left({\bf A}{\bf x} + \boldsymbol {\bf w}\right),
\end{equation}
where ${\bf y}\in \left\{+1, -1\right\}^{m}$,  ${\bf A}\in\mathbb{R}^{m\times n}$ is as above,
${\bf x} \in {\mathbb R}^n$ is the original signal, and $\boldsymbol {\bf w}
\in {\mathbb R}^m $ is additive white Gaussian noise with the variance  $\sigma^2$,
due to which some of the measurements signs may change with the respect to the noiseless measurements as given by  \eqref{1bitcs}.

\SubSection{Binary Iterative Hard Thresholding (BIHT)}
To recover ${\bf x}$ from ${\bf y}$, Jacques {\it et al} \cite{jacques2011robust} proposed the criterion
\begin{equation}
\label{BIHT}
\begin{split}
\min_{\bf x}\;  & f({\bf y}\odot {\bf A}{\bf x})\\
\mbox{subject\; to }\; & \left\|{\bf x} \right\|_2 = 1, \; {\bf x} \in \Sigma_K,
\end{split}
\end{equation}
where ``$\odot$" represents the Hadamard (element-wise) product,
${\Sigma_K} = \left\{{\bf x}\in \mathbb{R}^n : \left\|{\bf x}\right\|_0 \leq K \right\}$ (with
$\|{\bf v}\|_0$ denoting the number of non-zero components in ${\bf v}$)
is the set of $K$-sparse signals, and $f$ is one of the penalty functions
defined next. To penalize linearly the sign consistency violations, the choice is
$f({\bf z}) = 2\left\|{\bf z}_-\right\|_1$, where ${\bf z}_- = \min \left({\bf z}, 0\right)$ (where the minimum is
applied component-wise and the factor 2 is included for later convenience) and $\|{\bf v}\|_1 = \sum_i |v_i|$ is the
$\ell_1$ norm.  Quadratic penalization of the sign violations
is achieved by using  $f({\bf z})=\frac{1}{2} \left\|{\bf z}_-\right\|_2^2 $, where the factor $1/2$ is also
included for convenience. The {\it iterative hard thresholding} (IHT) \cite{IHT2009} algorithm applied to \eqref{BIHT}
(ignoring the norm constraint during the iterations) leads to the BIHT algorithm \cite{jacques2011robust}:
\vspace{0.1cm}
\begin{algorithm}{BIHT}{
\label{alg:BIHT}}
set $t =0, \tau >0, {\bf x}_0$ and $K$ \\
\qrepeat\\
     ${\bf v}_{t+1} = {\bf x}_{t} - \tau \partial f\left({\bf y}\odot {\bf A}{\bf x}_t\right)$\\
		 ${\bf x}_{t+1} =  {\mathcal P}_{\Sigma_K} \left( {\bf v}_{t+1}\right)$\\
		 $ t \leftarrow t+ 1$
\quntil some stopping criterion is satisfied.\\
\qreturn ${\bf x}_t/\left\|{\bf x}_t\right\|$
\end{algorithm}
\vspace{0.2cm}
In this algorithm, $\partial f$
denotes the subgradient of the objective (see \cite{jacques2011robust}, for details),
which is given by
\begin{equation}
\partial f\left({\bf y}\odot {\bf A}{\bf x}\right) = \left\{
\begin{array} {ll}
{\bf A}^T\left(\mbox{sign}({\bf A}{\bf x}) - {\bf y}\right),	& \mbox{$\ell_1$ penalty} \\
\left({\bf Y}{\bf A}\right)^T\left({\bf Y}{\bf A}{\bf x}\right)_-, & \mbox{$\ell_2$ penalty} ,
\end{array}\right.
\label{subgradient}
\end{equation}
where ${\bf Y} = \mbox{diag}({\bf y})$ is a diagonal matrix with vector ${\bf y}$ in its diagonal.
Step 3 corresponds to a sub-gradient descent step (with step-size $\tau$), while
Step 4 performs the projection onto the non-convex set ${\Sigma_K}$, which
corresponds to computing the best $K$-term approximation of ${\bf v}$, {\it i.e.},
keeping the $K$ largest components in magnitude and setting the others to zero. Finally, the
returned solution is projected onto the unit sphere to satisfy the constraint $\left\|{\bf x} \right\|_2 = 1$.
The versions of BIHT for the $\ell_1$ and $\ell_2$ penalties are referred to
as BIHT and BIHT-$\ell_2$, respectively.

\SubSection{Binary Fused Compressive Sensing (BFCS)}
We begin by introducing some notation.
The TV semi-norm of a vector ${\bf v} \in \mathbb{R}^n$ is given by
$\mbox{TV}({\bf v}) = \sum_{i=1}^{n-1} |v_{i+1} - v_i|$.
For $\varepsilon\geq 0$, we denote as $T_{\varepsilon}$ the
$\varepsilon$-radius TV ball, {\it i.e.}, $T_{\varepsilon} =
\left\{{\bf v}\in \mathbb{R}^p :\; \mbox{TV} \left({\bf v}\right)
 \leq \varepsilon \right\}$. The projection onto
$T_{\varepsilon}$ (denoted ${\mathcal P}_{T_{\varepsilon}}$) can be computed by the algorithm proposed
in \cite{fadili2011total}. Let $\mathfrak{G}({\bf v}) = \cup_{i=1}^{\mathcal{K}({\bf v})} \mathcal{G}_i({\bf v}) $,
where each $\mathcal{G}_k({\bf v}) \subset \{1,...,n\}$ is a set of consecutive indices
$\mathcal{G}_k({\bf v}) = \{i_k,...,i_k + |\mathcal{G}_k|-1\}$ such that, for $j\in\mathcal{G}_k$, $v_{j}\neq 0$, while
$v_{i_k-1}=0$ and $v_{i_k + |\mathcal{G}_k|} = 0$ (assume that $v_0 = v_{n+1} = 0$);
$\mathcal{G}_k({\bf v})$ is the $k$-th group of indices of consecutive
non-zero components of ${\bf v}$, and there are $\mathcal{K}({\bf v})$ such groups.
Let ${\bf v}_{\mathcal{G}_k} \in \mathbb{R}^{\left|\mathcal{G}_k\right|}$  be the sub-vector of
${\bf v}$ with indices in ${\mathcal{G}_k}$.


Obviously, the criterion in \eqref{BIHT} doesn't
encourage group-sparsity. To achieve that goal, we propose
the criterion
\begin{equation}\label{BFCS_SP2TV1}
\begin{split}
\min_{\bf x} \;&  f\left({\bf y}\odot {\bf A}{\bf x}\right)\\
\mbox{subject to} \; & \left\|{\bf x} \right\|_2 = 1, \; {\bf x} \in \Sigma_K \cap {S_{\epsilon}}
\end{split}
\end{equation}
where $S_{\epsilon}$ is defined as
\begin{equation} \label{eq:TVball}
 S_{\epsilon}= \left\{{\bf x}\in \mathbb{R}^n :\;  \overline{\mbox{TV}} \left({\bf x}_{\mathcal{G}_k}\right)
 \leq \epsilon, k = 1, \cdots, \mathcal{K}({\bf x}) \right\}.
\end{equation}
where
$\overline{\mbox{TV}} \left({\bf x}_{\mathcal{G}_k}\right) = \left( \left|{\mathcal{G}_k}\right|
 - 1\right)^{-1}\mbox{TV} \left({\bf x}_{\mathcal{G}_k}\right)$
 is a normalized TV, where
$ \left|{\mathcal{G}_k}\right| - 1$ is the number of absolute differences
in $\mbox{TV} \left({\bf x}_{\mathcal{G}_k}\right)$.
In contrast with a standard TV ball, $S_{\epsilon}$ promotes the ``fusion" of components only inside
each non-zero group, that is, the TV regularizer does not ``compete"  with the sparsity constraint
imposed by ${\bf x}\in\Sigma_K$.

To address the optimization problem in \eqref{BFCS_SP2TV1}, we propose the following
algorithm (which is a modification of BIHT):
\vspace{0.3cm}
\begin{algorithm}{BFCS}{
\label{alg:BFCS}}
set $t =0, \tau >0, \epsilon>0, K$ and ${\bf x}_0$ \\
\qrepeat\\
     ${\bf v}_{t+1} = {\bf x}_{t} - \tau \partial f\left({\bf y}\odot {\bf A}{\bf x}_t\right)$\\
		  ${\bf x}_{t+1} =  {\mathcal P}_{S_{\epsilon}}\! \left( {\mathcal P}_{\Sigma_K} ( {\bf v}_{t+1}) \right)$\\
		  $ t \leftarrow t+ 1$
\quntil some stopping criterion is satisfied.\\
\qreturn ${\bf x}_t/\left\|{\bf x}_t\right\|$
\end{algorithm}
\vspace{0.3cm}

Notice that the composition of projections in line 4 is not in general equal to the
projection on the non-convex set $\Sigma_K \cap S_{\epsilon}$, {\it i.e.}, ${\cal P}_{\Sigma_K \cap S_{\epsilon}}
\neq {\cal P}_{S_{\epsilon}} \circ {\cal P}_{\Sigma_K}$. However, this composition does
satisfy some relevant properties, which result from the structure of ${\cal P}_{S_{\epsilon}}$ expressed in
the following lemma (the proof of which is quite simple, but is omitted due to lack of space).

\vspace{0.2cm}
\begin{lemma}
\label{lem:lemma0}{\sl
Let ${\bf v} \in \mathbb{R}^n$ and ${\bf x}={\mathcal P}_{S_{\epsilon}} \left({\bf v}\right)$, then
\begin{equation}
\begin{split}
& {\bf x}_{\mathcal{G}_k}={\mathcal P}_{T_{\epsilon\left(\left|{\mathcal{G}_k}\right|-1\right)}} \left({\bf v}_{\mathcal{G}_k}\right),
\;\;\; \mbox{for}\;\; k = 1,\cdots, \mathcal{K}({\bf v}); \\
& {\bf x}_{ \overline{\mathfrak{G}}({\bf v}) } = {\bf 0},
\label{eq:projSP2TVball}
\end{split}
\end{equation}
where  $\overline{\mathfrak{G}}({\bf v}) = \left\{1,\cdots,n\right\} \setminus \mathfrak{G}({\bf v})$
and $\bf 0$ is a vector of zeros.}
\end{lemma}

The other relevant property of ${\cal P}_{S_{\epsilon}}$ is that it preserves sparsity, as expressed formally in the
following lemma.
\begin{lemma}
\label{lem:lemma1}{\sl
If ${\bf v} \in \Sigma_K$, then  ${\mathcal P}_{S_{\epsilon}}({\bf v}) \in \Sigma_K$.
Consequently, for any ${\bf v} \in \mathbb{R}^n$,
${\mathcal P}_{S_{\epsilon}}\! \left( {\mathcal P}_{\Sigma_K} ( {\bf v}) \right) \in \Sigma_K \cap S_{\epsilon}$.}
\end{lemma}

That is, although it is {\bf not} guaranteed that
${\cal P}_{S_{\epsilon}} \bigl( {\cal P}_{\Sigma_K}({\bf v})\bigr)$ coincides with the
orthogonal projection of ${\bf v}$ onto $\Sigma_K \cap S_{\epsilon}$, it belongs to this non-convex set.
In fact, the projection onto $\Sigma_K \cap S_{\epsilon}$ can be shown to be an NP-hard problem \cite{zeng2013binary},
since it belongs to the class of {\it shaped partition problems} \cite{hwang1999polynomial}, \cite{onn2004convex}
with variable number of parts.

\SubSection{Proposed Formulation and Algorithm}
In this paper we extend the BFCS approach to deal with the case where there
may exist some sign flips in the measurements. To this end, we adopt the
AOP technique \cite{yan2012robust}, yielding a new approach that we call
{\it robust BFCS} (RoBFCS); the similarly robust version of BIHT is termed
RoBIHT. Assume that there are at most $L$ sign flips and define the binary
vector $\boldsymbol{\Lambda} \in \{-1,+1\}^m$ as
 \begin{equation} \label{binaryvector}
\Lambda_i =\left\{ \begin{array}{lll}
-1  & \mbox{if} \; y_i \;\mbox{is} \; ``\mbox{flipped}'';\\
+1  & \mbox{otherwise.}
\end{array} \right.
\end{equation}
Then, the criterion of RoBFCS is given by
\begin{equation}\label{RobustBFCS}
\begin{split}
 \min_{{\bf x}\in \mathbb{R}^n,\; \boldsymbol{\Lambda}\in\{-1,+1\}^m } \;\;\;
 & f\left({\bf y}\odot \boldsymbol{\Lambda} \odot {\bf A}{\bf x}\right)\\
 \mbox{subject to}\; & \left\|{\bf x} \right\|_2 = 1, \; {\bf x} \in \Sigma_K \cap {S_{\epsilon}} \\
& \left\|\boldsymbol{\Lambda}_{-}\right\|_1 \leq L,
\end{split}
\end{equation}
where $\boldsymbol{\Lambda}_{-} = \min\{\boldsymbol{\Lambda}, {\bf 0}\}$. Problem
\eqref{RobustBFCS} is mixed continuous/discrete, and clearly difficult.
A natural approach to address \eqref{RobustBFCS} is via alternating minimization, as follows.
\vspace{0.3cm}
\begin{algorithm}{Framework of RoBFCS}{
\label{alg:BFCS}}
set $t =0, \boldsymbol{\Lambda}_0 = {\bf 1} \in \mathbb{R}^m, \epsilon>0, K, L$ and ${\bf x}_0$ \\
\qrepeat\\
     ${\bf x}_{t+1} = \Phi \left({\bf y}\odot\boldsymbol{\Lambda}_t, K, \epsilon\right)$\\
		  $\boldsymbol{\Lambda}_{t+1} = \Psi \left({\bf y}\odot {\bf A} {\bf x}_{t+1}, L\right)$\\
		  $ t \leftarrow t+ 1$
\quntil some stopping criterion is satisfied.\\
\qreturn ${\bf x}_t/\left\|{\bf x}_t\right\|$
\end{algorithm}
\vspace{0.3cm}

In this algorithm (template) lines 3 and 4 correspond to minimizing
\eqref{RobustBFCS} with respect to $\bf x$ and $\boldsymbol{\Lambda}$,
respectively.
The minimization w.r.t. ${\bf x}$ defines the function
\begin{equation}\label{RobustBFCS_1}
\begin{split}
\Phi \left( {\bf u}, K, \epsilon\right) = &
\arg\min_{{\bf x}\in \mathbb{R}^n } \;  f\left( {\bf u} \odot {\bf A}{\bf x}\right)\\
  &\mbox{subject to}\;  \left\|{\bf x} \right\|_2 = 1, \;  {\bf x} \in \Sigma_K \cap {S_{\epsilon}}
\end{split}
\end{equation}
which is an instance of \eqref{BFCS_SP2TV1}.
The minimization w.r.t. $\boldsymbol{\Lambda}$ defines the function
\begin{equation}\label{RobustBFCS_2}
\begin{split}
\Psi \left( {\bf z}, L\right) = & \arg\min_{\boldsymbol{\Lambda}\in \{-1,1\}^m } \;
f\left({\bf z}\odot \boldsymbol{\Lambda} \right)\\
 & \mbox{subject to} \; \left\|\boldsymbol{\Lambda}_{-}\right\|_1 \leq L
\end{split}
\end{equation}
As shown in \cite{yan2012robust}, \cite{movahed2012robust}, function
\eqref{RobustBFCS_2} is given in closed form by
\begin{equation} \label{binaryvector_update}
\bigl( \Psi \left( {\bf z}, L\right) \bigr)_i =\left\{ \begin{array}{lll}
-1  & \mbox{if} \; z_i \geq \tau;\\
+ 1  & \mbox{otherwise,}
\end{array} \right.
\end{equation}
where $\tau$ is the $L$-th largest element (in magnitude) of ${\bf z}$.

In the proposed RoBFCS algorithm, rather than implementing \eqref{RobustBFCS_1}
by running the BFCS algorithm until some stopping criterion is
satisfied, a single step thereof is applied, followed by the
implementation of \eqref{RobustBFCS_2} given by
\eqref{binaryvector_update}.

\vspace{0.3cm}
\begin{algorithm}{RoBFCS}{
\label{alg:BFCS}}
set $t =0, \tau >0, \epsilon>0, K, L$ and ${\bf x}_0, \boldsymbol{\Lambda}_0 = {\bf 1} \in \mathbb{R}^m$ \\
\qrepeat\\
     ${\bf v}_{t+1} = {\bf x}_{k} - \tau \partial f\left({\bf y}\odot \boldsymbol{\Lambda}_t \odot {\bf A}{\bf x}_k\right)$\\
		  ${\bf x}_{t+1} =  {\mathcal P}_{S_{\epsilon}} \left( {\mathcal P}_{\Sigma_K} ( {\bf v}_{t+1}) \right)$\\
			$\boldsymbol{\Lambda}_{t+1} = \Psi \left( {\bf y}\odot {\bf A} {\bf x}_{t+1}, L\right)$\\
		  $ t \leftarrow t + 1$
\quntil some stopping criterion is satisfied.\\
\qreturn ${\bf x}_t/\left\|{\bf x}_t\right\|$
\end{algorithm}
\vspace{0.3cm}	
The subgradient in line 3 is as given by \eqref{subgradient}, with ${\bf y}$ replaced with ${\bf y}\odot \boldsymbol{\Lambda}_t$.
If the original signal is known to be non-negative, the algorithm includes
a projection onto $\mathbb{R}_+^n$ in each iteration,
{\it i.e.}, line 4 becomes ${\bf x}_{k+1} = {\mathcal P}_{\mathbb{R}_+^n}\left( {\mathcal P}_{S_{\epsilon}}
\left( {\mathcal P}_{\Sigma_K} ( {\bf v}_{k+1}) \right)\right)$. The versions of RoBFCS (RoBIHT) with $\ell_1$ and
$\ell_2$ objectives are referred to as RoBFCS and RoBFCS-$\ell_2$ (RoBIHT and RoBIHT-$\ell_2$), respectively.

\Section{Experiments}
In this section, we report results of experiments aimed at studying the performance of RoBFCS.
All the experiments were performed using MATLAB on a 64-bit Windows 7 PC with an Intel Core i7 3.07 GHz
processor. In order to measure the performance of different algorithms, we employ the following
five metrics defined on an estimate ${\bf e}$ of an original vector ${\bf x}$ (both of unit norm):

\begin{itemize}
	\item Mean absolute error, $\textbf{MAE} = \left\|{\bf x} - {\bf e}\right\|_1/n$;
	\item Mean square error, $\textbf{MSE} = \left\|{\bf x} - {\bf e}\right\|_2^2/n$;
  \item Position error rate, $\textbf{PER}  = \sum_i \bigl\lvert |\mbox{sign}(x_i)| - |\mbox{sign}( e_i )  |  \bigr\rvert /n$;	
	\item Angle error, $\textbf{AE}  = \arccos\left\langle {\bf x}, {\bf e}\right\rangle / \pi$;
	\item Hamming error, $\textbf{HE} = \|\mbox{sign}({\bf A}{\bf x}) - \mbox{sign}({\bf A}{\bf e})\|_0 /m$.
\end{itemize}


\begin{table*}
\centering \caption{Experimental results} \label{tab:resultsofmetrics}
\begin{tabular}{|l|l|l|l|l|l|l|l|l|}
\hline
\footnotesize Metrics & \footnotesize BIHT    & \footnotesize BIHT-$\ell_2$ & \footnotesize BFCS    & \footnotesize BFCS-$\ell_2$ & \footnotesize RoBIHT  & \footnotesize RoBIHT-$\ell_2$ & \footnotesize RoBFCS  & \footnotesize RoBFCS-$\ell_2$ \\ \hline \hline

\footnotesize MAE     & \footnotesize 0.0019  & \footnotesize 0.0032        & \footnotesize 0.0008  & \footnotesize 0.0034        & \footnotesize 0.0019  & \footnotesize 0.0038          & \footnotesize \textbf{0.0001}  & \footnotesize 0.0038 \\
\footnotesize MSE     & \footnotesize 7.43E-5 & \footnotesize 1.65E-4       & \footnotesize 2.87E-5 & \footnotesize 1.78E-4       & \footnotesize 7.12E-5 & \footnotesize 2.04E-4         & \footnotesize \textbf{4.00E-7} & \footnotesize 2.06E-4 \\
\footnotesize PER     & \footnotesize 1.8\%   & \footnotesize 4.1\%         & \footnotesize 0.9\%   & \footnotesize 4.9\%         & \footnotesize 2.0\%   & \footnotesize 4.7\%           & \footnotesize \textbf{0\%} & \footnotesize 5.2\% \\
\footnotesize HE     & \footnotesize 0.0450  & \footnotesize 0.1360        & \footnotesize 0.0530  & \footnotesize 0.0995        & \footnotesize 0.0050  & \footnotesize 0.1420          & \footnotesize \textbf{0.0010}  & \footnotesize 0.1390\\
\footnotesize AE     & \footnotesize 0.1234  & \footnotesize 0.1852        & \footnotesize 0.0764  & \footnotesize 0.1927        & \footnotesize 0.1208  & \footnotesize 0.2070          & \footnotesize \textbf{0.0085}  & \footnotesize 0.2082 \\				
\hline
\end{tabular}
\end{table*}

The original signals ${\bf x}$ are taken as sparse and piece-wise smooth, of length $n = 2000$ with sparsity level
$K = 160$; specifically, \begin{equation} \label{piecewisesignal}
\bar{x}_i =\left\{ \begin{array}{lllll}
10 + 0.1 \, k_i, & i \in \cup_{i=1}^{0.25d}\mathcal{B}_i\\
15 + 0.1 \, k_i, & i \in \cup_{i=0.25d+1}^{0.5d}\mathcal{B}_i \\
-10 + 0.1 \, k_i, & i \in \cup_{i=0.5d+1}^{0.75d}\mathcal{B}_i \\
-15 + 0.1 \, k_i, & i \in \cup_{i=0.75d+1}^{d}\mathcal{B}_i \\
0, & i \not \in \cup_{i=1}^{d}\mathcal{B}_i
\end{array} \right.
\end{equation}
where the $k_i$ are independent samples of a zero-mean, unit variance Gaussian random variable,
$d$ is the number of non-zero groups of $\bf x$, and $\mathcal{B}_i,i \in \left\{1,\cdots,d\right\}$
indexes the $i$-th group, defined as
\[
\mathcal{B}_i = \left\{ 50+\left(i-1\right) n/d+1, \cdots, 50+\left(i-1\right)n/d+
K/d \right\}.
\]
The signal is then normalized, ${\bf x} = \bar{\bf x}/\|\bar{\bf x}\|_2$. The sensing matrix
${\bf A}$ is a $2000 \times 2000$ matrix with components sampled from the standard normal distribution.
 Finally, observations ${\bf y}$ are obtained by \eqref{1bitcsnoisy}, with noise standard deviation $\sigma = 1$.
The assumed number of sign flips is $L = 10$.

We run the algorithms BIHT, BIHT-$\ell_2$, BFCS, BFCS-$\ell_2$, RoBIHT, RoBIHT-$\ell_2$, RoBFCS and RoBFCS-$\ell_2$.
 The stopping criterion is  $\left\|{\bf x}_{(k+1)}-
{\bf x}_{(k)}\right\|/\left\|{\bf x}_{(k+1)}\right\|\leq 0.001$, where ${\bf x}_{(k)}$ is estimate
at the $k$-th iteration. Following the setup of \cite{jacques2011robust} and \cite{yan2012robust}, the step-size
of BIHT and RoBIHT and that of  BIHT-$\ell_2$ and Ro BIHT-$\ell_2$ is $\tau = 1$ and $1/m$, respectively.
While in BFCS, BFCS-$\ell_2$, RoBFCS, RoBFCS-$\ell_2$, $\tau$ and $\epsilon$ are hand tuned
for the best improvement in SNR. The quantitative results are shown in Table \ref{tab:resultsofmetrics}.

From Table \ref{tab:resultsofmetrics}, we can see that
RoBFCS performs the best in terms of the metrics considered.
Moreover, the algorithms with $\ell_1$ barrier perform better
than those with $\ell_2$ barrier.

\Section{Conclusions}
Based on the previously proposed  BFCS ({\it binary fused compressive sensing})
method, we have proposed an algorithm for recovering  sparse piece-wise smooth signals from
1-bit compressive measurements with some sign flips.
We have shown that if the original signals are in fact sparse and piece-wise smooth
and there are some sign flips in the measurements, the proposed method (termed RoBFCS --
{\it robust BFCS}) outperforms (under several accuracy measures)
the previous methods BFCS and BIHT ({\it binary iterative hard thresholding}). Future work will aim at making
RoBFCS adaptive in terms of $K$ and $L$.

\bibliographystyle{IEEEbib}
\bibliography{bibfile}

\end{document}